\documentclass[10pt,twocolumn,letterpaper]{article}
 
\usepackage{cvpr}
\usepackage{times}
\usepackage{epsfig}
\usepackage{graphicx}
\usepackage{amsmath}
\usepackage{amssymb}
\usepackage{arydshln}
\usepackage{bm}
\usepackage{mathtools}
\usepackage{graphicx}
\usepackage[export]{adjustbox}
\usepackage{caption}
\usepackage{subcaption}
\usepackage[pagebackref=true,breaklinks=true,letterpaper=true,colorlinks,bookmarks=false]{hyperref}

\cvprfinalcopy 

\newcommand{\ours}{D2D\xspace}
\newcommand{\oursfullname}{Describe-to-Detect\xspace}

\newcommand{\sota}{state-of-the-art\xspace}

\newcommand{\sas}{$\bm{S}_{\text{AS}}$\xspace}
\newcommand{\srs}{$\bm{S}_{\text{RS}}$\xspace}
\newcommand{\skpt}{$\bm{S}_{\text{D2D}}$\xspace}

\newtheorem{definition}{Definition}
\newtheorem{assumption}{Assumption}

\DeclarePairedDelimiter{\floor}{\lfloor}{\rfloor}

\def\cvprPaperID{4368} 

\ifcvprfinal\fi
\begin{document}

\title{D2D: Keypoint Extraction with Describe to Detect Approach}

\author{Yurun Tian$^{1}$ \hspace{6pt}Vassileios Balntas$^{2}$ \hspace{6pt}Tony Ng$^{1}$
\hspace{6pt}Axel Barroso-Laguna$^{1}$\\
\hspace{6pt}Yiannis Demiris$^{1}$ Krystian Mikolajczyk$^{1}$\\
$^1$ Imperial College London \\
$^2$ Scape Technologies\\
{\tt\small \{y.tian, tony.ng14, axel.barroso17, y.demiris, k.mikolajczyk\}@imperial.ac.uk}\\
{\tt\small vassileios@scape.io}}


\maketitle

\begin{abstract}
In this paper, we present a novel approach that exploits the information within the descriptor space to propose keypoint locations. Detect then describe, or detect and describe jointly are two typical strategies for extracting local descriptors. 
In contrast, we propose an approach that inverts this process by first describing and then detecting the keypoint locations. %
Describe-to-Detect (D2D) leverages successful descriptor models without the need for any additional training. Our method selects keypoints as salient locations with high information content which is defined by the descriptors rather than some independent operators.
We perform experiments on multiple benchmarks including image matching, camera localisation, and 3D reconstruction. The results indicate that our method improves the matching performance of various descriptors and that it generalises across methods and tasks.  

\end{abstract}


\section{Introduction}
\begin{figure}[!ht]
\centering
\includegraphics[trim={0cm 0cm 0cm 0cm}, width=\columnwidth]{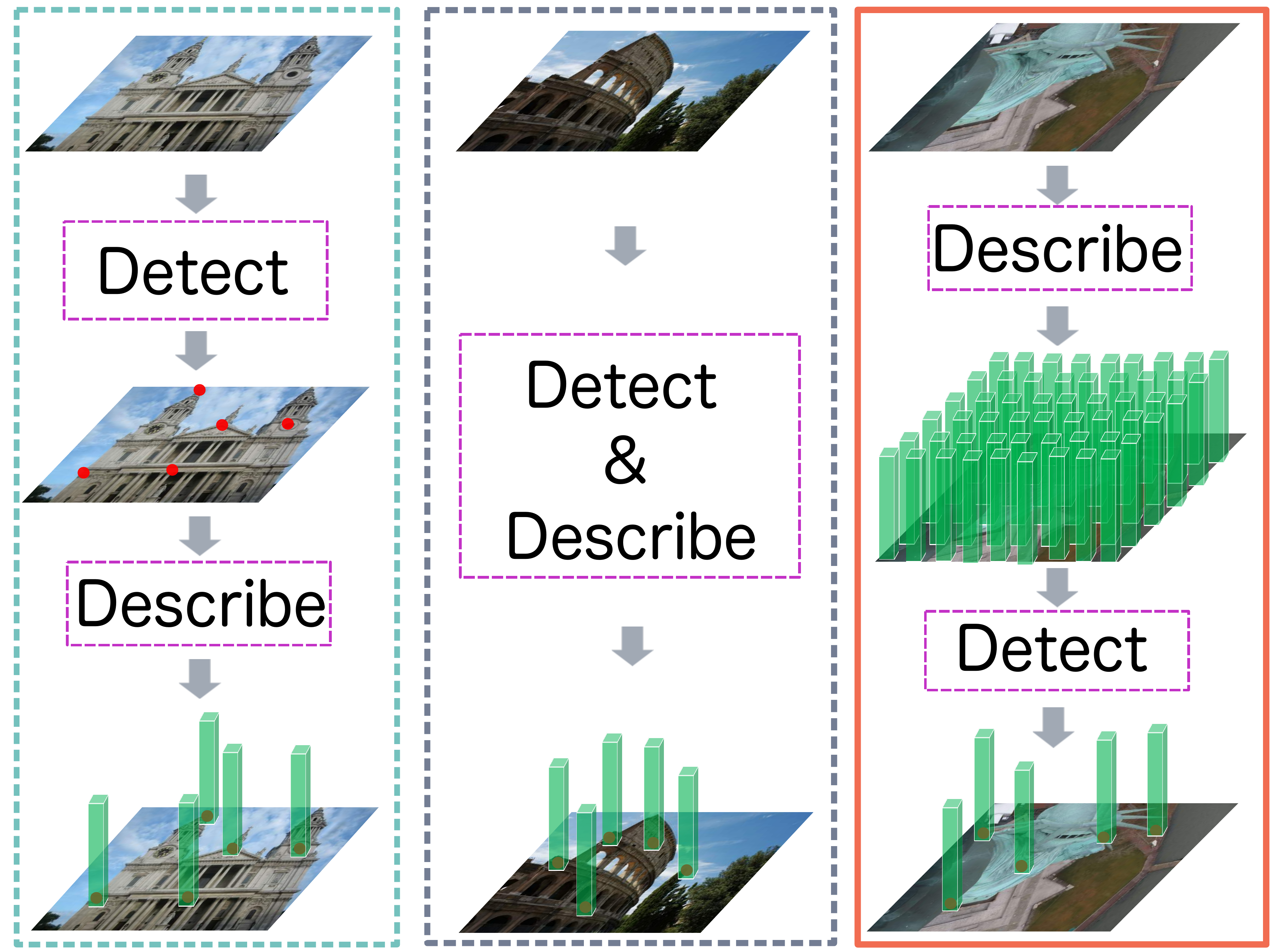}
\caption{Comparison of our proposed Describe-to-Detect framework~(right) to the existing Detect-then-Describe and Detect-and-Describe frameworks. }
\label{fig:teaser}
\end{figure}

One of the main problems in computer vision is concerned with the extraction of `meaningful' descriptions from images and sequences. These descriptions are then used for the correspondence problem which is critical for applications such as SLAM~\cite{monoslam2007,orbslam2015}, structure from motion~\cite{colmapcvpr2016,colmapeccv2016}, retrieval~\cite{gem2018}, camera localisation~\cite{sattler6dof}, tracking~\cite{visualtracking2013}, etc. The key issue is how to measure the `meaningfulness' from the data and which descriptors are the best. Extensive survey of salient region detectors~\cite{review_detect2008} attempts to identify the main properties expected from `good' features which include repeatability, informativeness, locality, quantity, accuracy, and efficiency. It has also been noted that the detector should be adapted to the needs of the application, \ie, the data. 

In contrast to the significant progress on local descriptors achieved with neural networks, keypoint detectors enjoyed little success from using learning methods, with few notable exceptions \cite{superpoint2018,d2net2019,keynet2019}.
As a consequence, keypoint detectors based on handcrafted filters such as Difference-of-Gaussians, Harris, Hessian~\cite{review_detect2008}, which all originate from research in 1980-ties are still used in many applications.





In the era of deep learning, there are three main research directions towards improving image matching, namely non-detector-specific description~\cite{l2net2017,hardnet2017,doap2018,sosnet2019}, non-descriptor-specific detection~\cite{tilde2015,keynet2019}, as well as jointly learnt detection-description~\cite{lift2016,lfnet2018,rfnet2019,d2net2019,r2d22019}.
What underlines the concept of disjoint frameworks is their sub-optimal compatibility between  the detection and description. In contrast to the CNN based descriptors \cite{l2net2017,hardnet2017,geodesc2018,doap2018,contextdesc2019,sosnet2019}, the performance of jointly learnt detection-description~\cite{lift2016,lfnet2018,d2net2019,r2d22019} does not seem to generalise well across datasets and tasks. 
CNN descriptors perform significantly better if trained and applied in the same data domain. Similarly, different keypoint detectors are suitable for different tasks. 
In addition, fine-tuning a descriptor for a specific keypoint detector further improves the performance. With all available options finding optimal pair of detector-descriptor for a dataset or a  task requires extensive experiments. 
Therefore, an approach that adapts keypoint detector to a descriptor without training and evaluation is highly sought for various applications.
%
%

Our approach is inspired by detectors based on various saliency measures \cite{kadir2001saliency,schiele2000recognition} where the saliency was defined in terms of local signal complexity or unpredictability; more specifically the Shannon entropy of local descriptor was suggested. Despite the appealing idea, such methods failed to be widely adopted due to the complexity of the required dense local measurements. However, currently available CNN dense descriptors allow revisiting the idea of using saliency measured by descriptors to define keypoint locations. 
Top performing learnt descriptors~\cite{l2net2017,hardnet2017,doap2018,sosnet2019} all share the same fully convolutional network (FCN) that adapts to varying image resolution and output dense descriptors.
Furthermore, joint methods like SuperPoint~\cite{superpoint2018}, D2-Net~\cite{d2net2019} and R2D2~\cite{r2d22019} also provide dense features.
%
%
%
%
The proposed approach can be seen as a combination of the classical saliency-based methods~\cite{kadir2001saliency,schiele2000recognition}
and the modern deep attention mechanisms~\cite{delf2017,gem2018,d2net2019}.

In summary, our main contributions are: 
\begin{itemize}
\item
We propose a novel \oursfullname (\ours) framework for keypoint detection that requires no training and adapts to any existing CNN based descriptor. 
\item
We propose a relative and an absolute saliency measure of local deep feature maps along the spatial and depth dimensions to define keypoints.
\item
We demonstrate  on several  benchmarks and different tasks that matching performance of various descriptors can be consistently improved by our approach.
\end{itemize}

\begin{figure*}[!ht]
\centering
\includegraphics[trim={0cm 0cm 0cm 0cm}, width=\linewidth]{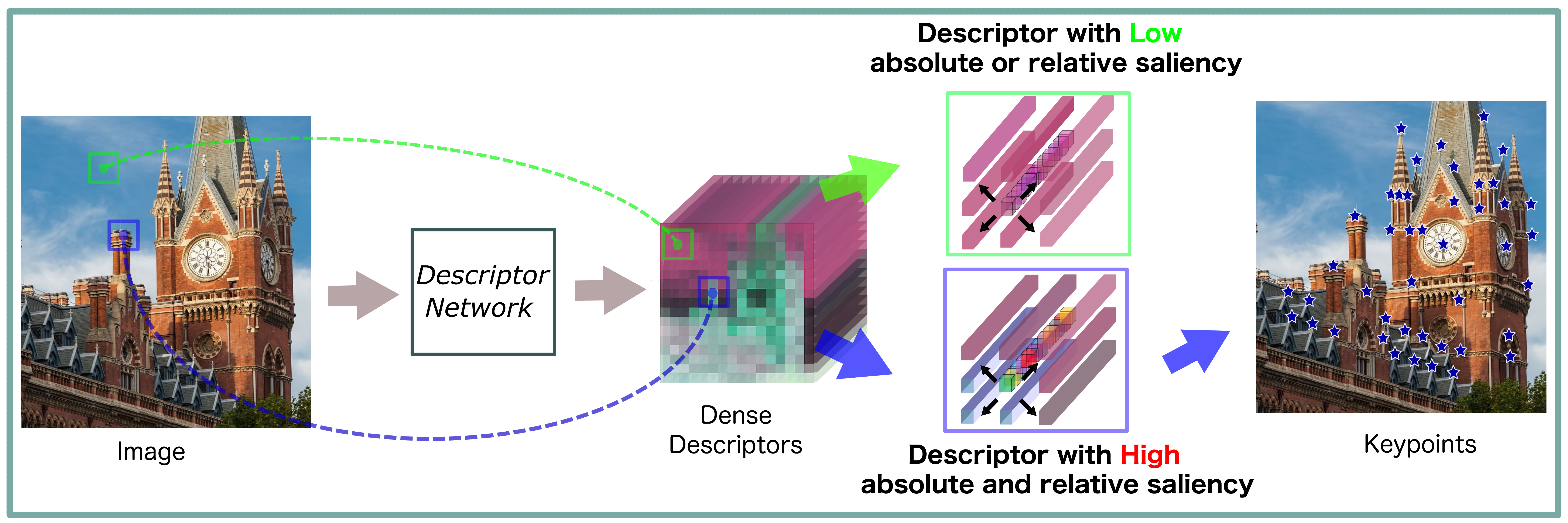}
\caption{The \oursfullname pipeline. Locations with high variation across channels (high absolute saliency) as well as high saliency w.r.t spatial neighbours (relative saliency) are detected as keypoints.}
\label{fig:pipeline}
\end{figure*}

\section{Related Works}
In this section, we briefly introduce some of the most recent learning-based methods for local feature detection and description. There are several survey articles that provide comprehensive reviews of this field \cite{review_desc2005,review_detect2005,review_detect2008,evaldetector2018}. 

\noindent \textbf{Local Feature Detection}.
Most of the existing hand-crafted~\cite{sift2004,brisk2011,kaze2012,saddle2016} or learned~\cite{tilde2015,codet2016,tcdet2017,quadnet2017,textdet2018,affnet2018,keynet2019} detectors are not descriptor-specific.
The main property required from keypoints is their repeatability such that their descriptors can be correctly matched.
TILDE~\cite{tilde2015} trains a  piece-wise linear regression model as the detector that is robust to weather and illumination changes.  CNN models are trained  with feature covariant constraints in \cite{codet2016,tcdet2017}.
Unsupervised trained 
QuadNet~\cite{quadnet2017} assumes that the ranking of the keypoint scores should be invariant to image transformations.
A similar idea is also explored in~\cite{textdet2018}
to detect keypoint in textured images.
AffNet~\cite{affnet2018} learns to predict the affine parameters of a local feature via the hard negative-constant loss based on the descriptors.
Key.Net~\cite{keynet2019} combines hand-crafted filters with learned ones to extract keypoints at different scale levels. Recently, it has been shown that pre-trained CNNs on standard tasks such as classification can be adapted to keypoint detection~\cite{elf2019}. However, the local feature matching pipeline is by nature different from classification.
In contrast, our method directly leverage CNNs pre-trained for description to achieve detection. 

\noindent \textbf{Local Feature Description}. 
The emergence of several large scale local patch datasets~\cite{ubc2011,hpatches2017,ps2018} stimulated the development of deep local descriptors~\cite{deepdesc2015,tfeat2016,l2net2017,hardnet2017,scaleaware2018,doap2018,sosnet2019} that are independent of the detectors. However, this paper is concerned with keypoint detection. Therefore we refer the reader to \cite{hpatches2017} for a detailed review and evaluation of recent descriptors.
In our experiments we include several recent descriptors such as HardNet \cite{hardnet2017}
and SOSNet \cite{sosnet2019}.
SIFT \cite{sift2004} is the most widely used handcrafted descriptor still considered as a well-performing baseline.
HardNet~\cite{hardnet2017} combines triplet loss with a within-batch hard negative mining that has proven to be remarkably effective and
SOSNet~\cite{sosnet2019} extends HardNet with and second-order loss.

\noindent \textbf{Joint Detection and Description}.
Joint training of detection-description has received more attention recently~\cite{lift2016,lfnet2018,superpoint2018,d2net2019,contextdesc2019,r2d22019,aslfeat2020,s2dnet2020}. 
SuperPoint~\cite{superpoint2018}, D2-Net~\cite{d2net2019}, and R2D2~\cite{r2d22019} are the three representatives of recent research direction, where patch cropping is replaced by fully convolutional dense descriptors.
SuperPoint~\cite{superpoint2018} leverages two separate decoders for  detection and description on a shared encoder.
Synthetic shapes and image pairs generated from random homographies are used to train the two parts.
In D2-Net~\cite{d2net2019}, local-maxima within and across channels of deep feature maps are defined as keypoints, with the same maps used for descriptors.
R2D2~\cite{r2d22019} aims at learning keypoints that are not only repeatable but also reliable together with robust descriptors. 
However, the computational cost for current joint frameworks is still high.
Besides, the generation of training data is typically laborious and method-specific.

Therefore, a keypoint detection method that is based on a trained descriptor model, thus adapted to the data without requiring any training, can be considered a novel and significant contribution.



\section{Describe-to-Detect}
In this section, we first define keypoints in terms of the descriptor saliency, then we present our approach to integrate \ours with existing \sota methods.

\subsection{What is a keypoint?}
Despite the absence of a unified definition, it is widely accepted that keypoints should be image points that have the potential of being repeatably detected under different imaging conditions.
As mentioned, according to \cite{review_detect2008}, such points should satisfy several requirements such as repeatability, informativeness, locality, quantity, accuracy and
efficiency.

In this work, we argue  that the informativeness, which we refer to as saliency, is the property that can lead to  satisfying most of the other requirements. 
We define the saliency in relative terms \ie w.r.t the other descriptors in the neighbourhood, as well as in  absolute terms as the information content of the descriptor.
Therefore, we depart from the following assumptions:
\begin{assumption}\label{assu:as}
\noindent A point in an image has a high absolute saliency if its corresponding descriptor is highly informative.
\end{assumption}

The idea of exploiting salient regions in an image has been adopted by many classical \cite{schiele2000recognition,kadir2001saliency} methods as well as recent attention-based models \cite{delf2017,gem2018,d2net2019}.
In tasks such as image retrieval, saliency/attention is defined on image regions with rich semantic information~\cite{gem2018,delf2017}.
In  feature matching, local image structures that exhibit significant variations in shape and texture can be considered salient.
However, absolute saliency alone is not sufficient for identifying keypoints.
For instance, highly informative but spatially non-discriminative structures should be avoided as they cannot be uniquely and accurately localised.
Therefore a relative saliency should also be considered.

\begin{assumption}\label{assu:rs}
\noindent A point in an image has a high relative saliency if its corresponding descriptor is highly discriminative in its spatial neighbourhood. 
\end{assumption}

The success of handcrafted detectors that define keypoints according to this criteria~\cite{moravec1980,sift2004,brisk2011,kaze2012,saddle2016,keynet2019} validates this assumption.
Descriptors on repeated textures can lead to geometrically noisy correspondences, therefore their  spatial uniqueness is essential. Similarly to the absolute saliency, the relative saliency alone is  not sufficient for detection.
For example, corner points of uniform regions can exhibit high relative saliency, whereas their descriptors information content is not high.

Based on Assumptions \ref{assu:as} and \ref{assu:rs}, our definition for keypoints based on  their corresponding descriptors is:
\begin{definition}\label{def:kpt}
\noindent A point in an image is a keypoint, if its corresponding descriptor's absolute and relative saliencies are both high.
\end{definition}

Definition \ref{def:kpt} is a generalization of the keypoints defined for low-level pixel intensities, either by simple operators such as autocorrelation \cite{moravec1980} or by early saliency based methods \cite{schiele2000recognition,kadir2001saliency}, to high-level descriptors.
In contrast to existing Detect-then/and-Describe frameworks, in Definition \ref{def:kpt}, we define the detector by the properties of the descriptor.
Thus, the key idea of  \oursfullname(\ours) is a description-guided detection.
Moreover, we claim that descriptors that are specifically trained to be robust to the changes of imaging conditions can provide data driven discriminativeness and thus, more reliable detections. 
It is worth noting that  our \ours  differs from other works that utilize the deep feature map response, but do not exploit the full representation potential of a descriptor.
For example, the detection step of D2-Net~\cite{d2net2019} is performed by considering each feature activation separately, as a score map for keypoint detection, whereas \ours detects keypoints via descriptor similarity in the metric space and therefore makes use of the rich information content across entire depth.

In summary, to identify the keypoints, Definition \ref{def:kpt} is concerned  with two properties: Firstly, when evaluating itself, the descriptor should be informative.
Secondly, when comparing to others, the descriptor should be discriminative.

\subsection{How to detect a keypoint?}

\noindent \textbf{Measuring the absolute saliency of a point} can be achieved by computing the entropy of a descriptor.
It has been shown in the design of binary descriptors~\cite{orb2011,bold2015}, that selecting binary tests with high entropy will encourage compact and robust representation. 
Therefore, we propose to measure the informativeness of a descriptor by its entropy, interpreted as a N-dimensional random variable. 
Unlike in binary descriptors where discrete entropy can be computed directly,  for real-valued descriptors differential entropy is needed.
However, computing an accurate differential entropy requires probability density estimation, which is computationally expensive.
Thus, similarly to the binary case~\cite{orb2011,bold2015}, we employ the standard deviation as a proxy for the entropy:
\begin{equation}\label{eq:as}
\mathbf{S}_{\text{AS}}(x,y) = \sqrt{\mathbb{E}[\bm{F}^2(x,y)]-\Bar{F}(x,y)^2}，
\end{equation}
where $\Bar{F}(x,y)$ is the mean value of descriptor $\bm{F}(x,y)$ across its dimensions.

\noindent \textbf{Measuring the relative saliency of a point} is based on Assumption \ref{assu:rs}.
A function that measures the relationship between a variable's current value and its neighboring values is the autocorrelation.
It has been successfully used by the classic Moravec corner detector~\cite{moravec1980} as well as the well known Harris detector \cite{harris1988}. However, their simple operators rely directly on pixel intensities which suffer from poor robustness to varying imaging conditions.
The autocorrelation was implemented as a sum of squared differences~(SSD) between the corresponding pixels of two overlapping patches:
\begin{equation}\label{eq:ssd}
\bm{S}_{\text{SSD}}(x,y) = \sum_{u}\sum_{v}\bm{W}(u,v)(\bm{I}(x,y)-\bm{I}(x+u,y+v))^2，
\end{equation}
where $I(x,y)$ indicate pixel intensity at  $(x,y)$, $(u,v)$ are window indexes centered at $(x,y)$, and $\bm{W}(u,v)$ are weights. A high value of $\bm{S}_{\text{SSD}}(x,y)$ means low similarity.
%
%
As a natural generalization of SSD for measuring the relative saliency, we  replace pixel intensities with dense descriptors :
\begin{equation}\label{eq:rs}
\bm{S}_{\text{RS}}(x,y) = \sum_{u}\sum_{v}\bm{W}(u,v)||\bm{F}(x,y)-\bm{F}(x+u,y+v)||_2，
\end{equation}
where  $\bm{F}(x,y)$ indicates the descriptor centered at location $(x,y)$, and $||\cdot||_2$ is the L2 distance.
A high value of $\bm{S}_{\text{RS}}(x,y)$ defines points with high relative saliency, \ie, this point stands out  from its neighbours according to the description provided by the pre-trained descriptor model.
Using Equations (\ref{eq:as}) and (\ref{eq:rs}), we assign a score to each point by: 
\begin{equation}\label{eq:key}
\mathbf{S}_{\text{D2D}}(x,y) = \mathbf{S}_{\text{AS}}(x,y) \mathbf{S}_{\text{RS}}(x,y).
\end{equation}


\subsection{Dense Descriptors}
All existing description methods can extract dense descriptors for a given image. 
For example, patch-based methods can be used to generate dense descriptors by extracting patches with a sliding window.
However, such strategy is infeasible in large scale tasks such as 3D reconstruction, due to its computational cost.
Fortunately, most recent \sota methods adopt the fully convolutional network architecture without fully-connected layers~\cite{l2net2017,hardnet2017,doap2018,sosnet2019,superpoint2018,d2net2019}.
Dense descriptor maps can be extracted with a single forward pass for images with various resolutions.
To guarantee the efficiency, we apply the proposed \ours to fully convolutional network descriptors only.
Specifically, in Section \ref{sec:exp}, we evaluate \ours with two \sota  descriptors, \ie, HardNet~\cite{hardnet2017} and SOSNet~\cite{sosnet2019}.  We further validate \ours on joint detection-description methods  SuperPoint~\cite{superpoint2018} and D2-Net~\cite{d2net2019}.

\subsection{Implementation Details}
\noindent \textbf{Computation of} $\bm{S}_{\text{AS}}(x,y)$ is done on descriptors before L2 normalization, since it has an effect of reducing the standard deviation magnitude across the dimensions.
It has been shown~\cite{hpatches2017} that the descriptor norm, that also reflects the magnitude of variance, is not helpful in the matching stage, however, we use it during the detection to identify informative points.

\noindent \textbf{Computation of} $\bm{S}_{\text{RS}}(x,y)$. 
We define the size of the window $\bm{W}(u,v)$ in Equation (\ref{eq:rs}) as $r_{\text{RS}}$. 
Considering that the receptive fields of neighbouring descriptors overlap and that the descriptor map resolution is typically lower than the input image, we sample the neighbouring descriptors with a step size of 2 and calculate the relative saliency with respect to the center descriptor. 
Note that the operation in Equation (\ref{eq:rs}) can be implemented efficiently with a convolution, therefore when the window size $r_{\text{RS}}$ is small and the sampling step is 2, the computational cost is negligible.    

\begin{figure}[t]
\centering
\includegraphics[trim={0cm 0cm 0cm 0cm}, width=\columnwidth]{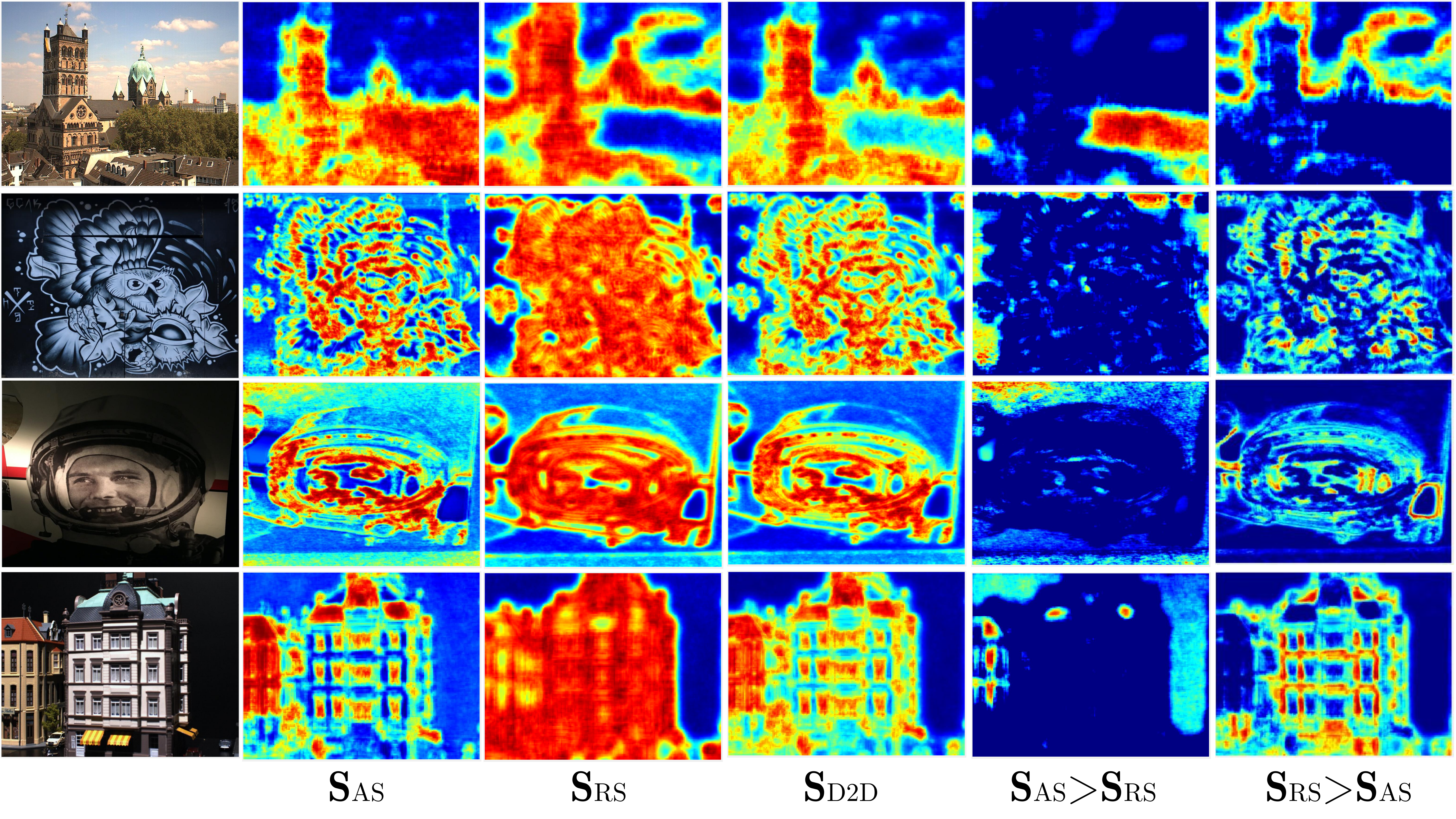}
\caption{Visualization of the heat maps generated by D2D applied to HardNet~\cite{hardnet2017}. From left to right the columns show images, heat maps of $\mathbf{S}_{\text{AS}}$, $\mathbf{S}_{\text{RS}}$, $\mathbf{S}_{\text{D2D}}$, $\max(0, \mathbf{S}_{\text{AS}} -\mathbf{S}_{\text{RS}})$ and $\max(0, \mathbf{S}_{\text{RS}}- \mathbf{S}_{\text{AS}})$, respectively. $\mathbf{S}_{\text{AS}}$ and $\mathbf{S}_{\text{RS}}$ are normalized so that their values are in $[0, 1]$.}
\label{fig:heatmaps}
\end{figure}

\begin{figure*}[ht!]
\centering
\includegraphics[trim={0cm 0cm 0cm 0cm}, width=\textwidth]{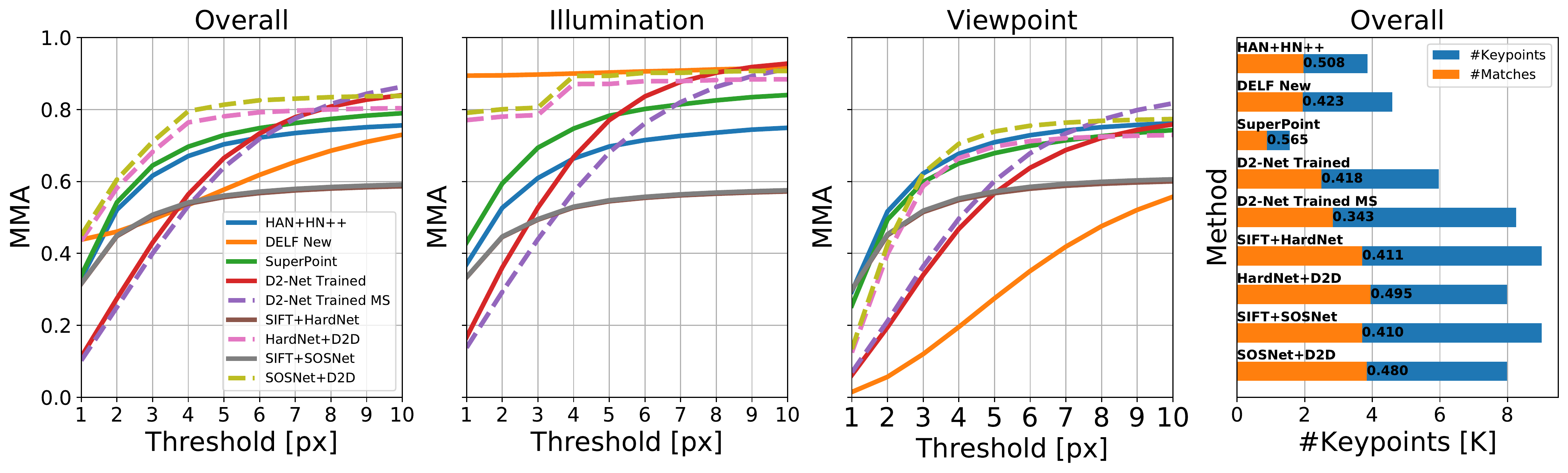}
\caption{Experimental results for the HPatches~\cite{hpatches2017} dataset. The results are reported with Mean Matching Accuracy. We observe that the proposed D2D method significantly outperforms other approaches, especially in the crucial high-accuracy area of $<5px$ }
\label{fig:hp_MMA}
\end{figure*}

\noindent \textbf{Combining \ours with  descriptors}.
To evaluate \ours we employ two current \sota patch-based descriptors, namely HardNet~\cite{hardnet2017} and SOSNet~\cite{sosnet2019}.
Given the network architecture~\cite{l2net2017} and an input image of size $H$ $\times$ $W$($H \ge 32, W \ge 32$), the output feature map size is $(\floor*{H/4}-7)$ $\times$ $(\floor*{W/4}-7)$. 
The receptive field is of size $51$ $\times$ $51$.
Therefore, each descriptor $\bm{F}(x,y)$ describes a $51$ $\times$ $51$ region centered at $(4x+14, 4y+14)$.
There are two stride-$2$ convolutional layers in the network, therefore $\bm{F}$ describes each $51$ $\times$ $51$ patch with stride of 4.
In other words, keypoints are at least 4 pixels away from each other.
Such sparse sampling has also been validated in other works~\cite{r2d22019,unsuperpoint2019}. %
Finally, with $\mathbf{S}_{\text{D2D}}$ we directly take the top $K$ ranked points as keypoints.

In D2-Net~\cite{d2net2019}, the effect of downsampling layers is mitigated by upsampling the dense descriptors.
However, with a large receptive overlap, dense $\bm{F}$ is redundant.
For example, $\bm{F}(x,y)$ and $\bm{F}(x+1,y)$ describe two $51$ $\times$ $51$ patch with a $47$ $\times$ $51$ overlap.
For networks such as HardNet~\cite{hardnet2017} and SOSNet~\cite{sosnet2019} that are trained to be insensitive to such small changes, additional interpolation of feature maps is unnecessary. 

Also, note that the amount of content the network can see in a $51$ $\times$ $51$ region is defined by the resolution of the image. High resolution and dense sampling can make the neighbouring descriptors indistinguishable.
An interesting question is whether  a multi-scale strategy to tackle the scale changes is needed.
We show in Section~\ref{sec:exp} that single scale HardNet~\cite{hardnet2017} and SOSNet~\cite{sosnet2019} perform well in different tasks.
We claim that there are two reasons for this:
First, dramatic scale changes are rare in typical images of the same scenes. 
Second, scale changes are often global and the ranking of the detected keypoints is not affected by such changes\cite{quadnet2017}.

Furthermore, we give some examples in Figure~\ref{fig:heatmaps} to show different components of the final keypoint score map and how \sas and \srs contribute to \skpt.
As shown, \sas highlights all regions that have high intensity variations, while \srs has high scores in structured areas.
Finally, \skpt combines the two parts, resulting in a low score for repeated/non-textured areas and edges.
Points with \srs greater than \sas are informative but not locally discriminative.
This includes repeated textures like tree leaves and tiles on building roof, as well as intensity noise in visually homogeneous regions.
Otherwise, line structures are less informative but can be discriminative from the adjacent regions, which results in  \sas greater than \srs.

\section{Experiments}
\label{sec:exp}
In this section we present the results for various tasks on different datasets, which include image matching,  visual localisation and 3D reconstruction.

\subsection{Comparison with the \sota}
We evaluate \ours on three different tasks, \ie, image matching, visual localisation, and 3D reconstruction on three standard benchmarks, \ie, Hpatches~\cite{hpatches2017}, Aachen Day-Night~\cite{sattler2012image,sattler2018benchmarking}, and ETH SfM~\cite{eth_benchmark2017}, respectively.
Each of the tasks tests the compatibility of the detector and the descriptor from a different perspective.
We employ HardNet and SOSNet trained on Liberty from UBC dataset~\cite{ubc2011}. 
For all experiments in this section, we set $r_{\text{RS}}$ to be 5.

We evaluate our method on three different tasks, \ie, image matching, visual localisation, and 3D reconstruction on three standard benchmarks, \ie, Hpatches~\cite{hpatches2017}, Aachen Day-Night~\cite{sattler2012image,sattler2018benchmarking}, and ETH SfM~\cite{eth_benchmark2017}.
Each of the tasks tests the compatibility of the detector and the descriptor from a different perspective.

\noindent \textbf{Image Matching.}
Hpatches~\cite{hpatches2017} dataset contains 116 image sequences with ground truth homographies under different viewpoint or illumination changes.
Following the evaluation protocol of~\cite{d2net2019,review_desc2005}, we report the mean matching accuracy (MMA).
In Figure~\ref{fig:hp_MMA}, we report MMA for thresholds $1$ to $10$ pixels averaged over all image pairs.
Also, we give the mean number of keypoints, mean number of mutual nearest neighbour matches per image pair, and the ratio between the two numbers.

As shown, combining \ours approach with HardNet and SOSNet can achieve superior or comparable results to other \sota methods.
By comparing the curves of HardNet/SOSNet+\ours with SIFT+HardNet/SOSNet, we can observe that the \ours finds more compatible keypoints for  HardNet/SOSNet  than SIFT.
Also note that when using the SIFT detector, the MMA curves of HardNet and SOSNet almost overlap, however, \ours helps to further reveal their performance difference.
This also demonstrates that the detector is a very crucial component of matching, and that optimising descriptor independently from the detector is insufficient.
Moreover, we can also see that \ours can detect more keypoints thus leading to a higher number of mutual nearest neighbour matches, which beneficial for various applications.
Besides, HardNet+\ours also surpass AffNet+HardNet++, where AffNet is specifically trained with a descriptor loss.
This shows that leveraging the absolute and relative saliency of descriptors is an effective approach to detect keypoints.



\begin{table}
\footnotesize
\centering
\begin{tabular}{c@{\hspace{6pt}}c@{\hspace{4pt}}c@{\hspace{6pt}}c@{\hspace{6pt}}c@{\hspace{6pt}}c@{\hspace{6pt}}c@{\hspace{6pt}}}
\hline
Method & $\bf{\#}$\bf{Dim} & $\bf{\#}$\bf{Kp} &0.5m,~$2^\circ$ & 1m,~$5^\circ$ & 5m,~$10^\circ$ \\ \hline \hline
SIFT &128  &11K&33.7 & 52.0& 65.3\\
DELF(New)~\cite{delf2017}  &1024 &11K&39.8 & 61.2& 85.7\\
HAN+HN++~\cite{affnet2018,hardnet2017}         &128  &11K &39.8 &61.2 &77.6\\
SuperPoint~\cite{superpoint2018} &128  &3.7K &42.8&57.1&75.5\\
D2-Net SS~\cite{d2net2019} &512  &12K&44.9&66.3&88.8\\
D2-Net MS~\cite{d2net2019} &512  &12K&44.9&64.3&88.8\\
R2D2 (N=8)~\cite{delf2017}  &128 &10K &\textbf{45.9} &\textbf{66.3}& \textbf{88.8}\\

\hline
SIFT+HardNet~\cite{hardnet2017}         &128  &11K &34.7&52.0&69.4\\
HardNet+\ours      &128 &16K&\underline{41.8}&\underline{61.2}&\underline{84.7}\\ \hdashline

SIFT+SOSNet~\cite{sosnet2019}          &128  &11ßK &36.7&53.1&70.4\\
SOSNet+\ours       &128  &16K &\underline{42.9}&\underline{64.3}&\underline{85.7}\\\hdashline

\end{tabular}
\caption{Comparison to the state of the art on the Aachen Day-Night dataset. We report the percentages of successfully localized images within 3 error thresholds as in~\cite{d2net2019,r2d22019}.}
\label{tab:aachen}
\end{table}
\begin{table*}[htp]
\footnotesize
\begin{center}
\begin{tabular}{c l c c c c c c}
\hline

& & \bf{\# Image} & \bf{\# Registered} 
& \bf{\# Sparse Points} & \bf{\# Dense Points} & \bf{Track Length} & \bf{Reproj. Error}  \\
\hline  

\bf{Fountain} & SIFT & 11 & 11 & 14K & 292K & 4.79 & \textcolor{red}{0.39px}\\
& SuperPoint & & 11  &7K &304K & 4.93& 0.81px \\
& D2-Net & & 11  &19K &301K & 3.03& 1.40px \\
& HardNet+\ours & & 11 & \textcolor{red}{20K} & 304K & 6.27 & 1.34px\\

& SOSNet+\ours & & 11 & \textcolor{red}{20K} & \textcolor{red}{305K} & \textcolor{red}{6.41} & 1.36px\\
\hline

\bf{Herzjesu} & SIFT & 8 & 8 & 7K & 241K & 4.22 & \textcolor{red}{0.43px}\\
& SuperPoint & & 8  & 5K & \textcolor{red}{244K}& 4.47& 0.79px \\
& D2-Net & & 8  & 13K & 221K& 2.87& 1.37px \\
& HardNet+\ours & & 8 & \textcolor{red}{13K} &242K & 5.73 & 1.29px\\
& SOSNet+\ours & & 8 & \textcolor{red}{13K} & 237K & \textcolor{red}{6.06} & 1.34px\\
\hline

\bf{South Building} & SIFT & 128 & 128 & 108K &\textcolor{red}{2.14M} & 6.04 & \textcolor{red}{0.54px}\\
& SuperPoint & & 128  & 125k & 2.13M& 7.10& 0.83px \\
& D2-Net & & 128  &178K &2.06M &3.11 & 1.36px \\
& HardNet+\ours & & 128 & \textcolor{red}{193K} & 2.02M & 8.71 & 1.33px\\
& SOSNet+\ours & & 128 & 184K & 1.94M & \textcolor{red}{8.99} & 1.36px\\
\hline

\bf{Madrid Metropolis} & SIFT & 1344 & 500 & 116K & \textcolor{red}{1.82M} & \textcolor{red}{6.32} & \textcolor{red}{0.60px}\\
& SuperPoint & &702  & 125K & 1.14M & 4.43& 1.05px \\
& D2-Net & & 787  &229K & 0.96M& 5.50& 1.27px \\
& HardNet+\ours & &\textcolor{red}{899} & \textcolor{red}{710K} &1.13M &5.31 & 1.08px\\

& SOSNet+\ours & & 865 & 626K & 1.15M & 6.00 & 1.14px\\
\hline

\bf{Gendarmenmarkt} &SIFT &1463 &1035  &338K &\textcolor{red}{4.22M} &5.52 & \textcolor{red}{0.69px}\\
& SuperPoint & &1112  &236K & 2.49M&4.74 & 1.10px \\
& D2-Net & & 1225  & 541K& 2.60M& 5.21& 1.30px \\
& HardNet+\ours & &1250 & \textcolor{red}{1716K} &2.64M &5.32 & 1.16px\\
& SOSNet+\ours & & \textcolor{red}{1255} & 1562K & 2.71M & \textcolor{red}{5.95} & 1.20px\\
\hline

\hline
\end{tabular}
\end{center}
\normalsize
\caption{Evaluation results on ETH dataset ~\cite{eth_benchmark2017} for SfM. We can observe that with our proposed \ours, the shallow networks trained on local patches can significantly surpass deeper ones trained on larger datasets with full resolution images.}
\label{tab:eth_3d_recon}
\end{table*}

\noindent \textbf{Day-Night Visual Localisation.}
In this section, we further evaluate our method on the task of long-term visual localization using the Aachen Day-Night dataset~\cite{sattler2018benchmarking,sattler2012image}.
This task evaluates the performance of local features under challenging conditions including day-night and viewpoint changes.
Our evaluation is performed via a localisation pipeline\footnote{\scriptsize \url{{https://github.com/tsattler/visuallocalizationbenchmark/tree/master/local_feature_evaluation}}} based on COLMAP \cite{colmapcvpr2016} and The Visual Localization Benchmark\footnote{\scriptsize \url{https://www.visuallocalization.net/}}.

In Table~\ref{tab:aachen}, we report the percentages of successfully localized images within three error thresholds.
As can be seen, \ours significantly boost the performance of HardNet and SOSNet.
Even though D2-Net and R2D2 are still the best performers on this dataset,  their advantage may come from the training data or network architecture, \ie, D2-Net uses VGG16 network~\cite{vgg162014} pre-trained on ImageNet and then trained on MegaDepth~\cite{megadepth2018} while R2D2 is also trained on Aachen Day-Night dataset.
However, HardNet and SOSNet are only trained on 450K $32 \times 32$ patches from Liberty dataset~\cite{ubc2011}.
We will show in the next experiments that, these two models trained on patches labeled by an SfM pipeline are especially effective for 3D reconstruction tasks.

\noindent \textbf{3D Reconstruction.}
We test our method on the ETH SfM benchmark~\cite{eth_benchmark2017} in the task of 3D reconstruction.
We compare the reconstruction quality by comparing the number of registered images, reconstructed sparse and dense points, mean track length, and the reprojection error.
Following~\cite{eth_benchmark2017}, no nearest neighbour ratio test is conducted to better expose the matching performance of the descriptors.
The reconstruction results are listed in Table~\ref{tab:eth_3d_recon}. As shown, HardNet/SOSNet+\ours shows consistent performance increase in terms of the number of registered images, the number of sparse points, and the track length, which are important indicators of the reconstruction quality.
This observation is expected as in this experiment, both HardNet and SOSNet are trained on local patches that are extracted and labeled via the SfM pipeline, and therefore are more suitable for this task.

\begin{figure}[ht]
\centering
\includegraphics[trim={0cm 0cm 0cm 0cm}, width=\columnwidth]{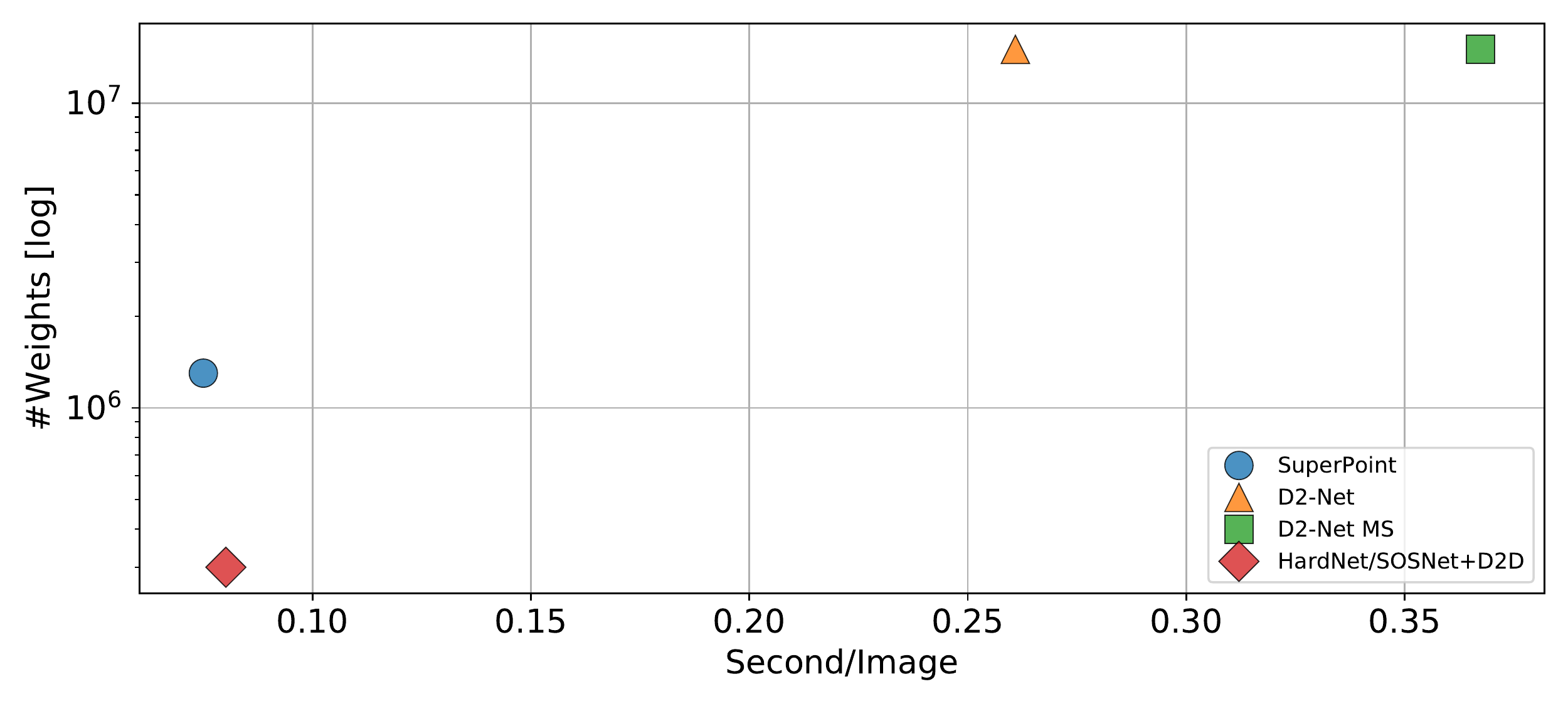}
\caption{Comparison of efficiency.}
\label{fig:speed}
\vspace{-10pt}
\end{figure}

\noindent \textbf{Efficiency}.
In this experiment, we compare the feature extraction speed of several methods. 
Specifically, we record the extraction time over 108 image sequences in Hpatches~\cite{hpatches2017}, where there are 648 images with various resolutions (the average resolution is $775 \times 978$). All methods are tested on a RTX 2080 GPU, and the results are shown in Figure~\ref{fig:speed}.
SuperPoint and D2-Net has 1.3M and 15M parameters, respectively, whereas HardNet/SOSNet+\ours only relies 0.3M.
Worth noting that R2D2 also uses similar architecture to HardNet/SOSNet, however it has no downsampling layers, thus the computational cost increases linearly with the depth.
HardNet/SOSNet+\ours is slightly slower than SuperPoint, due to the extra time that is mostly spend on ranking the $\mathbf{S}_{\text{D2D}}$ score of keypoints, whereas SuperPoint takes a thresholding operation. 

In summary, from the results on three different tasks with three different datasets we observe that with \ours, patch-based descriptors HardNet and SOSNet can achieve competitive performance compared to joint detection-description methods such as D2-Net and SuperPoint.
With significantly less parameters and faster speed, HardNet and SOSNet can achieve comparable/superior results to/than the \sota methods.
These results validate our hypothesis that the networks trained for descriptors can be also used for detection.

\subsection{Ablation Study}


\begin{table}
\footnotesize
\centering
\begin{tabular}{|c@{\hspace{6pt}}|c@{\hspace{8pt}}||c@{\hspace{6pt}}|c@{\hspace{6pt}}|c@{\hspace{6pt}}|c@{\hspace{6pt}}|}
\hline
AS &  RS  & SuperPoint&D2-Net & HardNet& SOSNet\\ \hline
$\surd$&        &67.51 &61.20 &71.38 &72.66 \\ \hline
      &$\surd$ &67.58 &60.07 &69.32 &72.77 \\ \hline
$\surd$&$\surd$ &\textbf{67.64} &\textbf{61.42} &\textbf{72.40}  &\textbf{75.40} \\  \hline
\end{tabular}
\caption{Ablative study in terms of Absolute Saliency(AS) and Relative Saliency(RS). Numbers are in terms of the average MMA on Hpatches~\cite{hpatches2017} across pixel error threshold 1 to 10.}
\label{tab:as_rs}
\end{table}

\noindent \textbf{Combining D2D with joint methods}.
In order to further validate the effectiveness of the proposed \ours, we test it in combination with detect-and-describe methods\footnote{By the time of submission, there was no released code or model available for R2D2~\cite{r2d22019}, therefore we omitted it.} namely D2-Net\cite{d2net2019} and SuperPoint~\cite{superpoint2018}.
Each of the two methods has its unique detection strategy:
SuperPoint detects via thresholding of  deep score maps while D2-Net selects local maxima.
We adapt \ours in the following way:
For SuperPoint, we generate a new threshhold $\alpha^{*}$ by:
\begin{equation}\label{eq:alpha_sp}
\alpha^{*}  = \frac{\mathbb{E}[\bm{S}_{\text{D2D}}\bm{S}_{\text{O}}]}{\mathbb{E}[\bm{S}_{\text{O}}]}\alpha,
\end{equation}
where $\alpha$ and $\bm{S}_{\text{O}}$ are the original threshold and score map, respectively.
For D2-Net, we choose local maxima that also have high $\mathbf{S}_{\text{D2D}}$.
Specifically, if $(x,y)$ is a keypoint than it should be detected by the non-maxima-suppression as well as have:
\begin{equation}\label{eq:d2}
 \bm{S}_{\text{D2D}}(x,y)  > \mathbb{E}[\bm{S}_{\text{D2D}}]
\end{equation}

\begin{figure}[!htb]
\centering
\includegraphics[trim={0cm 0cm 0cm 0cm}, width=\columnwidth]{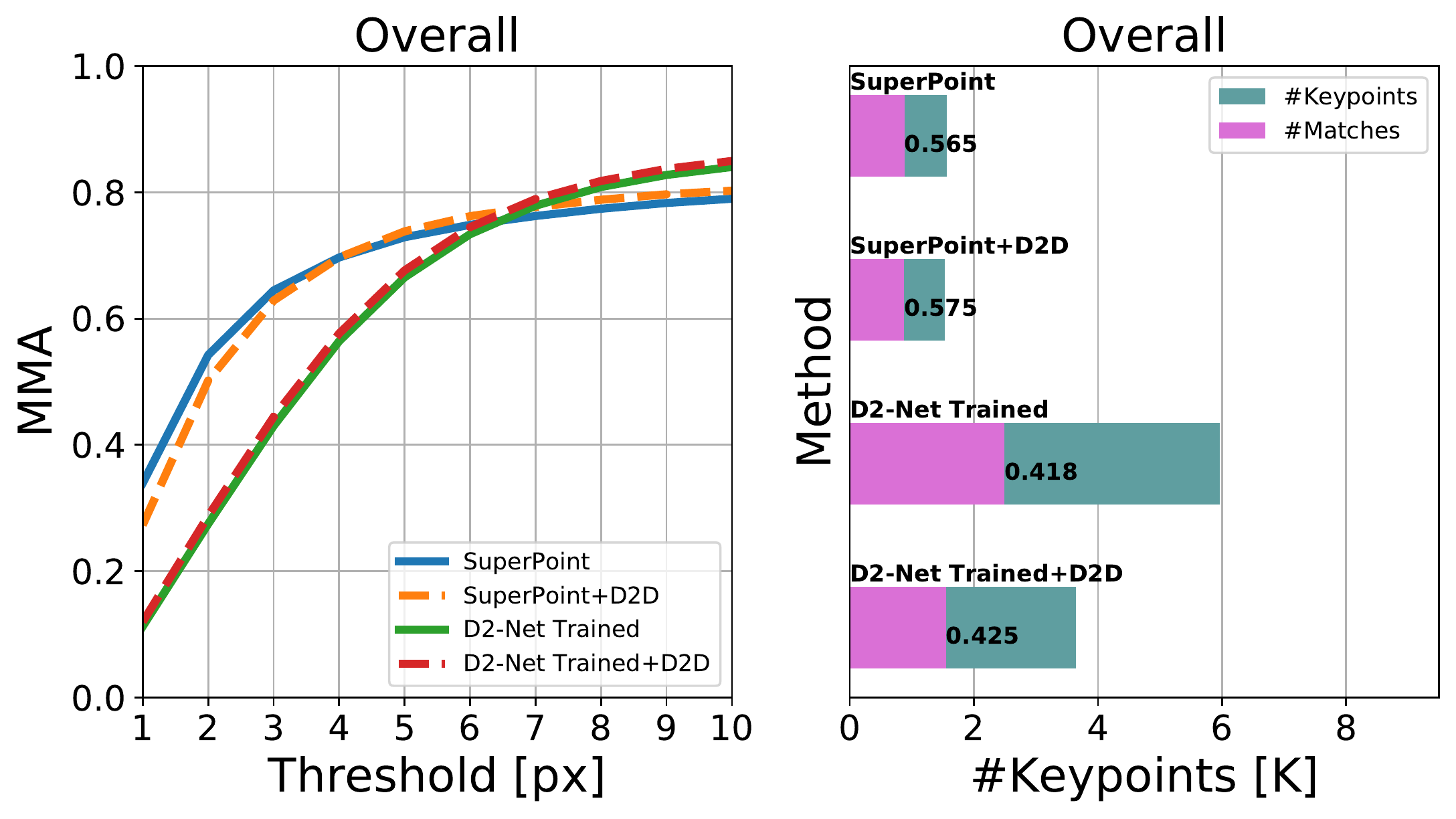}
\caption{Performance of combining \ours with SuperPoint~\cite{superpoint2018} and D2-Net~\cite{d2net2019} on Hpatches~\cite{hpatches2017}.}
\label{fig:hp_jointmethods}
\end{figure}

In Figure~\ref{fig:hp_jointmethods}, \ours  improves the MMA score and the ratio of mutual nearest neighbour matches on Hpatches~\cite{hpatches2017}. 
Moreover, in Table~\ref{tab:aachen_jointmethods}, SuperPoint+\ours achieves remarkably better localisation accuracy.
 D2-Net+\ours can maintain the same accuracy with much fewer detections indicating that keypoints not contributing to the localisation are filtered out by \ours.
These results  demonstrates that \ours can also improve the jointly trained detection-description methods.

\begin{table}
\footnotesize
\centering
\begin{tabular}{c@{\hspace{1pt}}c@{\hspace{4pt}}c@{\hspace{6pt}}c@{\hspace{6pt}}c@{\hspace{6pt}}c@{\hspace{6pt}}c@{\hspace{6pt}}}
\hline
Method & $\bf{\#}$\bf{Dim} & $\bf{\#}$\bf{Kp} &0.5m,~$2^\circ$ & 1m,~$5^\circ$ & 5m,~$10^\circ$ \\ \hline \hline
SuperPoint &256  &3.7K &42.8&57.1&75.5\\
SuperPoint+\ours   &256  &3.7K &41.8 &\underline{59.2}&\underline{78.6}\\ \hdashline
D2-Net SS &512  &12K&44.9&66.3&88.8\\
D2-Net SS+\ours       &512  &\underline{8.3K} &\underline{44.9}& \underline{66.3}& \underline{88.8}\\\hdashline
\end{tabular}
\caption{Performance of combining \ours with SuperPoint~\cite{superpoint2018} and D2-Net~\cite{d2net2019} on Aachen Day-Night~\cite{sattler2012image,sattler2018benchmarking}}
\label{tab:aachen_jointmethods}
\end{table}

\noindent \textbf{Impact of absolute and relative descriptor saliency}.
In Table~\ref{tab:as_rs}, we show how \sas and \srs impact the matching performance.
We observe that each of the two terms  enables the detection, and the performance is further boosted when they are combined.
This indicates that the absolute and relative saliency, \ie, informativeness and distinctiveness  of a point are two effective and complementary factors. 

\begin{figure}[ht]
\centering
\includegraphics[trim={0cm 0cm 0cm 0cm}, width=\columnwidth]{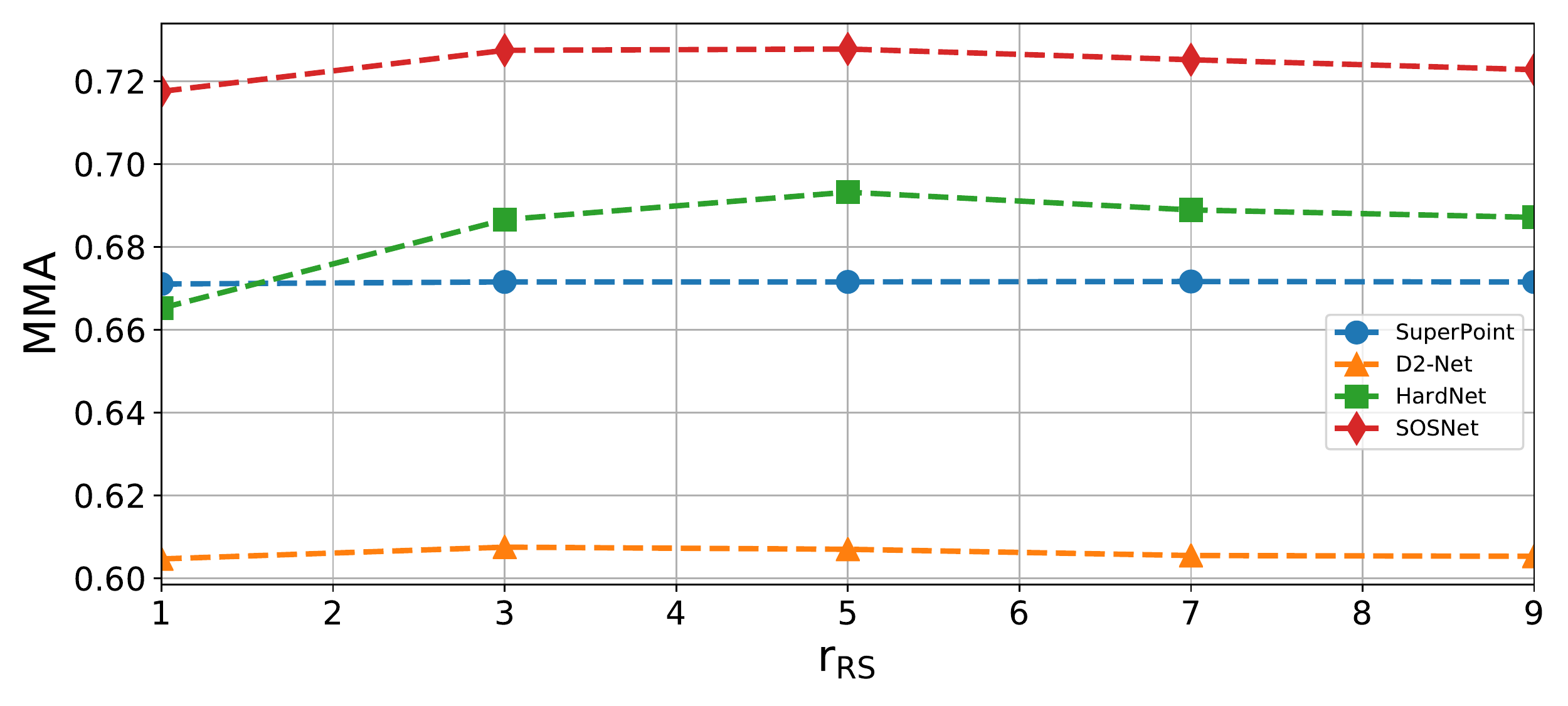}
\caption{Performance in terms of MMA with different choice of $r_{\text{RS}}$.}
\label{fig:r_rs}
\end{figure}

\noindent \textbf{Impact of $r_{\text{RS}}$}.
Matching performance in terms of different window size $r_{\text{RS}}$ for computing relative saliency  is shown in Figure~\ref{fig:r_rs}, where the experiment is done using only \srs as the keypoint score.
For HardNet and SOSNet, the best $r_{\text{RS}}$ is 5, which means that it is better to compare patches that are 20 pixels (stride 4 times 5) away from the center, which is approximately half of the receptive field size. Descriptors that are too close are indistinguishable.   

\noindent \textbf{Keypoint complementarity.}
Table~\ref{tab:rep_diff_det} shows the results of a repeatability test across different descriptors combined with \ours. This is to demonstrate the complementarity  of keypoints detected with different methods.  The off diagonal scores are normalised with the diagonal scores  for example,  keypoints from HardNet+\ours are compared to those detected by  SOSNet+\ours. Low normalised repeatability score indicates that the keypoints are mostly different \ie different locations, thus the methods are complementary. Similarly HardNet and SOSNet give high score. This may be expected as both share the same architecture and similar training process.   However,  high repeatability between SuperPoint and D2-Net which indicates that the two descriptors are not complementary \ie measure the same type of information that \ours uses for detecting keypoints. 


\begin{table}[!htp]
\footnotesize
\centering
\begin{tabular}{|c|c@{\hspace{5pt}}|c@{\hspace{10pt}}|c@{\hspace{10pt}}|c@{\hspace{10pt}}|}
\hline
 & SuperPoint  &  D2-Net &  HardNet &  SOSNet\\ \hline
 SuperPoint&1&1.0154&0.745&0.765 \\ \hline
 D2-Net&1.136 &1 &0.675 &0.690 \\ \hline
 HardNet &0.849 &0.729 &1  &0.952 \\ \hline
 SOSNet &0.868 &0.738 &0.950 &1
 \\ \hline
\end{tabular}
\caption{Keypoint repeatbality on Hpatches~\cite{hpatches2017} with different detectors. Column:detector used on source image. Row:detector used on destination image. Numbers are the percentage of repeatbality change in terms of the original repeatbality~(diagonal).}
\label{tab:rep_diff_det}

\end{table}

\section{Conclusion}
We proposed a new \oursfullname (\ours) framework for the task of keypoint detection given dense descriptors. 
We have demonstrated that CNN models  trained to describe  can also be used to detect. \ours is simple, does not require training, is efficient and can be combined with any existing descriptor.
We defined the descriptor saliency as the most important property and  proposed an absolute and relative saliency measure to select keypoints that are highly informative in descriptor space and discriminative in their local spacial neighbourhood.

Our experimental evaluation  on three different tasks and different datastes show that \ours offers a significant boost to the matching performance of various descriptors. It also improves results for camera localisation and 3D reconstruction. 

\noindent{\bf Acknowledgements.} This work was supported by the UK EPSRC research grant EP/S032398/1.



{\small
\bibliographystyle{ieee}
\bibliography{egbib}

\begin{thebibliography}{10}\itemsep=-1pt

\bibitem{kaze2012}
P.~F. Alcantarilla, A.~Bartoli, and A.~J. Davison.
\newblock Kaze features.
\newblock In {\em European Conference on Computer Vision}, pages 214--227.
  Springer, 2012.

\bibitem{saddle2016}
J.~Aldana-Iuit, D.~Mishkin, O.~Chum, and J.~Matas.
\newblock In the saddle: chasing fast and repeatable features.
\newblock In {\em 2016 23rd International Conference on Pattern Recognition
  (ICPR)}, pages 675--680. IEEE, 2016.

\bibitem{hpatches2017}
V.~Balntas, K.~Lenc, A.~Vedaldi, and K.~Mikolajczyk.
\newblock Hpatches: A benchmark and evaluation of handcrafted and learned local
  descriptors.
\newblock In {\em Proceedings of the IEEE Conference on Computer Vision and
  Pattern Recognition (CVPR)}, volume~4, page~6, 2017.

\bibitem{tfeat2016}
V.~Balntas, E.~Riba, D.~Ponsa, and K.~Mikolajczyk.
\newblock Learning local feature descriptors with triplets and shallow
  convolutional neural networks.
\newblock In {\em British Machine Vision Conference (BMVC)}, volume~1, page~3,
  2016.

\bibitem{bold2015}
V.~Balntas, L.~Tang, and K.~Mikolajczyk.
\newblock Bold - binary online learned descriptor for efficient image matching.
\newblock In {\em The IEEE Conference on Computer Vision and Pattern
  Recognition (CVPR)}, June 2015.

\bibitem{elf2019}
A.~Benbihi, M.~Geist, and C.~Pradalier.
\newblock Elf: Embedded localisation of features in pre-trained cnn.
\newblock In {\em Proceedings of the IEEE International Conference on Computer
  Vision}, pages 7940--7949, 2019.

\bibitem{ubc2011}
M.~Brown, G.~Hua, and S.~Winder.
\newblock Discriminative learning of local image descriptors.
\newblock {\em IEEE PAMI}, 33(1):43--57, 2011.

\bibitem{unsuperpoint2019}
P.~H. Christiansen, M.~F. Kragh, Y.~Brodskiy, and H.~Karstoft.
\newblock Unsuperpoint: End-to-end unsupervised interest point detector and
  descriptor.
\newblock {\em arXiv preprint arXiv:1907.04011}, 2019.

\bibitem{monoslam2007}
A.~J. Davison, I.~D. Reid, N.~D. Molton, and O.~Stasse.
\newblock Monoslam: Real-time single camera slam.
\newblock {\em IEEE Transactions on Pattern Analysis \& Machine Intelligence},
  (6):1052--1067, 2007.

\bibitem{superpoint2018}
D.~DeTone, T.~Malisiewicz, and A.~Rabinovich.
\newblock Superpoint: Self-supervised interest point detection and description.
\newblock In {\em Proceedings of the IEEE Conference on Computer Vision and
  Pattern Recognition Workshops}, pages 224--236, 2018.

\bibitem{d2net2019}
M.~Dusmanu, I.~Rocco, T.~Pajdla, M.~Pollefeys, J.~Sivic, A.~Torii, and
  T.~Sattler.
\newblock D2-net: A trainable cnn for joint detection and description of local
  features.
\newblock {\em arXiv preprint arXiv:1905.03561}, 2019.

\bibitem{s2dnet2020}
H.~Germain, G.~Bourmaud, and V.~Lepetit.
\newblock S2dnet: Learning accurate correspondences for sparse-to-dense feature
  matching.
\newblock {\em arXiv preprint arXiv:2004.01673}, 2020.

\bibitem{harris1988}
C.~G. Harris, M.~Stephens, et~al.
\newblock A combined corner and edge detector.
\newblock In {\em Alvey vision conference}, volume~15, pages 10--5244.
  Citeseer, 1988.

\bibitem{doap2018}
K.~He, Y.~Lu, and S.~Sclaroff.
\newblock Local descriptors optimized for average precision.
\newblock In {\em Proceedings of the IEEE Conference on Computer Vision and
  Pattern Recognition (CVPR)}, pages 596--605, 2018.

\bibitem{kadir2001saliency}
T.~Kadir and M.~Brady.
\newblock Saliency, scale and image description.
\newblock {\em International Journal of Computer Vision}, 45(2):83--105, 2001.

\bibitem{scaleaware2018}
M.~Keller, Z.~Chen, F.~Maffra, P.~Schmuck, and M.~Chli.
\newblock Learning deep descriptors with scale-aware triplet networks.
\newblock In {\em Proceedings of the IEEE Conference on Computer Vision and
  Pattern Recognition (CVPR)}. IEEE, 2018.

\bibitem{keynet2019}
A.~B. Laguna, E.~Riba, D.~Ponsa, and K.~Mikolajczyk.
\newblock Key.net: Keypoint detection by handcrafted and learned cnn filters.
\newblock {\em arXiv preprint arXiv:1904.00889}, 2019.

\bibitem{codet2016}
K.~Lenc and A.~Vedaldi.
\newblock Learning covariant feature detectors.
\newblock In {\em European Conference on Computer Vision}, pages 100--117.
  Springer, 2016.

\bibitem{evaldetector2018}
K.~Lenc and A.~Vedaldi.
\newblock Large scale evaluation of local image feature detectors on homography
  datasets.
\newblock {\em arXiv preprint arXiv:1807.07939}, 2018.

\bibitem{brisk2011}
S.~Leutenegger, M.~Chli, and R.~Siegwart.
\newblock Brisk: Binary robust invariant scalable keypoints.
\newblock In {\em 2011 IEEE international conference on computer vision
  (ICCV)}, pages 2548--2555. Ieee, 2011.

\bibitem{megadepth2018}
Z.~Li and N.~Snavely.
\newblock Megadepth: Learning single-view depth prediction from internet
  photos.
\newblock In {\em Proceedings of the IEEE Conference on Computer Vision and
  Pattern Recognition}, pages 2041--2050, 2018.

\bibitem{sift2004}
D.~G. Lowe.
\newblock Distinctive image features from scale-invariant keypoints.
\newblock {\em Proceedings of the IEEE International Conference on Computer
  Vision (ICCV)}, 60(2):91--110, 2004.

\bibitem{contextdesc2019}
Z.~Luo, T.~Shen, L.~Zhou, J.~Zhang, Y.~Yao, S.~Li, T.~Fang, and L.~Quan.
\newblock Contextdesc: Local descriptor augmentation with cross-modality
  context.
\newblock In {\em Proceedings of the IEEE Conference on Computer Vision and
  Pattern Recognition}, pages 2527--2536, 2019.

\bibitem{geodesc2018}
Z.~Luo, T.~Shen, L.~Zhou, S.~Zhu, R.~Zhang, Y.~Yao, T.~Fang, and L.~Quan.
\newblock Geodesc: Learning local descriptors by integrating geometry
  constraints.
\newblock In {\em European Conference on Computer Vision (ECCV)}, pages
  170--185. Springer, 2018.

\bibitem{aslfeat2020}
Z.~Luo, L.~Zhou, X.~Bai, H.~Chen, J.~Zhang, Y.~Yao, S.~Li, T.~Fang, and
  L.~Quan.
\newblock Aslfeat: Learning local features of accurate shape and localization.
\newblock {\em arXiv preprint arXiv:2003.10071}, 2020.

\bibitem{review_desc2005}
K.~Mikolajczyk and C.~Schmid.
\newblock A performance evaluation of local descriptors.
\newblock {\em IEEE PAMI}, 27(10):1615--1630, 2005.

\bibitem{review_detect2005}
K.~Mikolajczyk, T.~Tuytelaars, C.~Schmid, A.~Zisserman, J.~Matas,
  F.~Schaffalitzky, T.~Kadir, and L.~Van~Gool.
\newblock A comparison of affine region detectors.
\newblock {\em International journal of computer vision}, 65(1-2):43--72, 2005.

\bibitem{hardnet2017}
A.~Mishchuk, D.~Mishkin, F.~Radenovic, and J.~Matas.
\newblock Working hard to know your neighbor's margins: Local descriptor
  learning loss.
\newblock In {\em Advances in Neural Information Processing Systems (NIPS)},
  pages 4826--4837, 2017.

\bibitem{affnet2018}
D.~Mishkin, F.~Radenovic, and J.~Matas.
\newblock Repeatability is not enough: Learning affine regions via
  discriminability.
\newblock In {\em The European Conference on Computer Vision (ECCV)}, September
  2018.

\bibitem{ps2018}
R.~Mitra, N.~Doiphode, U.~Gautam, S.~Narayan, S.~Ahmed, S.~Chandran, and
  A.~Jain.
\newblock A large dataset for improving patch matching.
\newblock {\em arXiv preprint arXiv:1801.01466}, 2018.

\bibitem{moravec1980}
H.~P. Moravec.
\newblock Obstacle avoidance and navigation in the real world by a seeing robot
  rover.
\newblock Technical report, Stanford Univ CA Dept of Computer Science, 1980.

\bibitem{orbslam2015}
R.~Mur-Artal, J.~M.~M. Montiel, and J.~D. Tardos.
\newblock Orb-slam: a versatile and accurate monocular slam system.
\newblock {\em IEEE transactions on robotics}, 31(5):1147--1163, 2015.

\bibitem{delf2017}
H.~Noh, A.~Araujo, J.~Sim, T.~Weyand, and B.~Han.
\newblock Large-scale image retrieval with attentive deep local features.
\newblock In {\em Proceedings of the IEEE International Conference on Computer
  Vision}, pages 3456--3465, 2017.

\bibitem{lfnet2018}
Y.~Ono, E.~Trulls, P.~Fua, and K.~M. Yi.
\newblock Lf-net: learning local features from images.
\newblock In {\em Advances in Neural Information Processing Systems}, pages
  6234--6244, 2018.

\bibitem{gem2018}
F.~Radenovi{\'c}, G.~Tolias, and O.~Chum.
\newblock Fine-tuning cnn image retrieval with no human annotation.
\newblock {\em IEEE transactions on pattern analysis and machine intelligence},
  41(7):1655--1668, 2018.

\bibitem{r2d22019}
J.~Revaud, P.~Weinzaepfel, C.~De~Souza, N.~Pion, G.~Csurka, Y.~Cabon, and
  M.~Humenberger.
\newblock R2d2: Repeatable and reliable detector and descriptor.
\newblock {\em arXiv preprint arXiv:1906.06195}, 2019.

\bibitem{orb2011}
E.~Rublee, V.~Rabaud, K.~Konolige, and G.~R. Bradski.
\newblock Orb: An efficient alternative to sift or surf.
\newblock In {\em ICCV}, volume~11, page~2. Citeseer, 2011.

\bibitem{sattler2018benchmarking}
T.~Sattler, W.~Maddern, C.~Toft, A.~Torii, L.~Hammarstrand, E.~Stenborg,
  D.~Safari, M.~Okutomi, M.~Pollefeys, J.~Sivic, et~al.
\newblock Benchmarking 6dof outdoor visual localization in changing conditions.
\newblock In {\em Proceedings of the IEEE Conference on Computer Vision and
  Pattern Recognition}, pages 8601--8610, 2018.

\bibitem{sattler6dof}
T.~Sattler, W.~Maddern, A.~Torii, J.~Sivic, T.~Pajdla, M.~Pollefeys, and
  M.~Okutomi.
\newblock Benchmarking 6dof urban visual localization in changing conditions.
\newblock {\em CoRR}, abs/1707.09092, 2017.

\bibitem{sattler2012image}
T.~Sattler, T.~Weyand, B.~Leibe, and L.~Kobbelt.
\newblock Image retrieval for image-based localization revisited.
\newblock In {\em BMVC}, volume~1, page~4, 2012.

\bibitem{quadnet2017}
N.~Savinov, A.~Seki, L.~Ladicky, T.~Sattler, and M.~Pollefeys.
\newblock Quad-networks: unsupervised learning to rank for interest point
  detection.
\newblock In {\em Proceedings of the IEEE conference on computer vision and
  pattern recognition}, pages 1822--1830, 2017.

\bibitem{schiele2000recognition}
B.~Schiele and J.~L. Crowley.
\newblock Recognition without correspondence using multidimensional receptive
  field histograms.
\newblock {\em International Journal of Computer Vision}, 36(1):31--50, 2000.

\bibitem{colmapcvpr2016}
J.~L. Schonberger and J.-M. Frahm.
\newblock Structure-from-motion revisited.
\newblock In {\em Proceedings of the IEEE Conference on Computer Vision and
  Pattern Recognition (CVPR)}, pages 4104--4113, 2016.

\bibitem{eth_benchmark2017}
J.~L. Sch\"{o}nberger, H.~Hardmeier, T.~Sattler, and M.~Pollefeys.
\newblock Comparative evaluation of hand-crafted and learned local features.
\newblock In {\em Proceedings of the IEEE Conference on Computer Vision and
  Pattern Recognition (CVPR)}, 2017.

\bibitem{colmapeccv2016}
J.~L. Sch{\"o}nberger, E.~Zheng, J.-M. Frahm, and M.~Pollefeys.
\newblock Pixelwise view selection for unstructured multi-view stereo.
\newblock In {\em European Conference on Computer Vision (ECCV)}, pages
  501--518. Springer, 2016.

\bibitem{rfnet2019}
X.~Shen, C.~Wang, X.~Li, Z.~Yu, J.~Li, C.~Wen, M.~Cheng, and Z.~He.
\newblock Rf-net: An end-to-end image matching network based on receptive
  field.
\newblock In {\em Proceedings of the IEEE Conference on Computer Vision and
  Pattern Recognition}, pages 8132--8140, 2019.

\bibitem{deepdesc2015}
E.~Simo-Serra, E.~Trulls, L.~Ferraz, I.~Kokkinos, P.~Fua, and F.~Moreno-Noguer.
\newblock Discriminative learning of deep convolutional feature point
  descriptors.
\newblock In {\em Proceedings of the IEEE International Conference on Computer
  Vision (ICCV)}.

\bibitem{vgg162014}
K.~Simonyan and A.~Zisserman.
\newblock Very deep convolutional networks for large-scale image recognition.
\newblock {\em arXiv preprint arXiv:1409.1556}, 2014.

\bibitem{visualtracking2013}
A.~W. Smeulders, D.~M. Chu, R.~Cucchiara, S.~Calderara, A.~Dehghan, and
  M.~Shah.
\newblock Visual tracking: An experimental survey.
\newblock {\em IEEE transactions on pattern analysis and machine intelligence},
  36(7):1442--1468, 2013.

\bibitem{l2net2017}
Y.~Tian, B.~Fan, F.~Wu, et~al.
\newblock L2-net: Deep learning of discriminative patch descriptor in euclidean
  space.
\newblock In {\em Proceedings of the IEEE Conference on Computer Vision and
  Pattern Recognition (CVPR)}, volume~1, page~6, 2017.

\bibitem{sosnet2019}
Y.~Tian, X.~Yu, B.~Fan, F.~Wu, H.~Heijnen, and V.~Balntas.
\newblock Sosnet: Second order similarity regularization for local descriptor
  learning.
\newblock In {\em Proceedings of the IEEE Conference on Computer Vision and
  Pattern Recognition}, pages 11016--11025, 2019.

\bibitem{review_detect2008}
T.~Tuytelaars, K.~Mikolajczyk, et~al.
\newblock Local invariant feature detectors: a survey.
\newblock {\em Foundations and trends{\textregistered} in computer graphics and
  vision}, 3(3):177--280, 2008.

\bibitem{tilde2015}
Y.~Verdie, K.~Yi, P.~Fua, and V.~Lepetit.
\newblock Tilde: a temporally invariant learned detector.
\newblock In {\em Proceedings of the IEEE Conference on Computer Vision and
  Pattern Recognition}, pages 5279--5288, 2015.

\bibitem{lift2016}
K.~M. Yi, E.~Trulls, V.~Lepetit, and P.~Fua.
\newblock Lift: Learned invariant feature transform.
\newblock In {\em European Conference on Computer Vision}, pages 467--483.
  Springer, 2016.

\bibitem{textdet2018}
L.~Zhang and S.~Rusinkiewicz.
\newblock Learning to detect features in texture images.
\newblock In {\em Proceedings of the IEEE Conference on Computer Vision and
  Pattern Recognition}, pages 6325--6333, 2018.

\bibitem{tcdet2017}
X.~Zhang, F.~X. Yu, S.~Karaman, and S.-F. Chang.
\newblock Learning discriminative and transformation covariant local feature
  detectors.
\newblock In {\em Proceedings of the IEEE Conference on Computer Vision and
  Pattern Recognition}, pages 6818--6826, 2017.

\end{thebibliography}
}

\end{document}